\documentclass[runningheads]{llncs}

\usepackage{eccv}

\usepackage{eccvabbrv}

\usepackage{graphicx}
\usepackage{booktabs}
\usepackage{multirow}
\usepackage{float}

\usepackage[accsupp]{axessibility}  

\usepackage{hyperref}

\usepackage{orcidlink}

\begin{document}

\title{WaterGen: Decoupling Scene and Medium in Underwater Image Generation} 


\author{Jiayi Wu\thanks{Equal contribution}\inst{1} \and
Tianfu Wang$^\star$\inst{1} \and
Tianyi Xiong\inst{1} \and
Dehao Yuan\inst{1} \and
Xiaomin Lin\inst{2} \and
Md Jahidul Islam\inst{3} \and
Cornelia Fermuller\inst{1} \and
Christopher Metzler\inst{1} \and
Yiannis Aloimonos\inst{1}}

\authorrunning{J.~Wu, T.~Wang et al.}

\institute{University of Maryland \and
University of South Florida \and
University of Florida}
\maketitle

\begin{abstract}

Underwater computer vision tasks, such as detection, restoration, and segmentation, are limited by the scarcity of large-scale and diverse training data. We introduce WaterGen, a method for generating large-scale, realistic, and diverse underwater images that provides independent control of the scene and water medium conditions. 
Our approach treats underwater image generation as the decoupled control of two factors: realistic and diverse scene content (what is in the image), and accurate and controllable water medium effects (what the water does to the image). 
Existing methods generally achieve only part of this objective: they either provide controllability with limited realism or diversity, or generate realistic scenes without accurately and independently modeling water-medium effects.
Our key insight, that allows us to avoid this compromise, is that scene generation and medium modeling can be decoupled within a latent diffusion framework, enabling diverse scene generation together with accurate and controllable underwater appearance. 
To do this, we decompose underwater image synthesis into two stages.
First, we fine-tune the latent diffusion U-Net using degradation-free underwater images so that it learns to generate diverse and realistic latent embeddings of underwater scene content without medium-induced degradation.
Second, we formulate the physically accurate medium degradation synthesis as a conditional decoding process applied to these latent embeddings.
This decoupled design allows our model to generate diverse scenes with full control of underwater appearance.
We leverage WaterGen to build large-scale synthetic underwater datasets that are diverse in scene structures and accurate in water effects and pseudo-labels. We demonstrate that our synthetic data consistently improve downstream performance in underwater restoration and semantic segmentation. Code and model weights are available at \url{https://github.com/jiayi-wu-umd/WaterGen}.

\end{abstract}

\section{Introduction}
\label{sec:intro}
\begin{figure}[t]
\centering
\includegraphics[width=\linewidth]{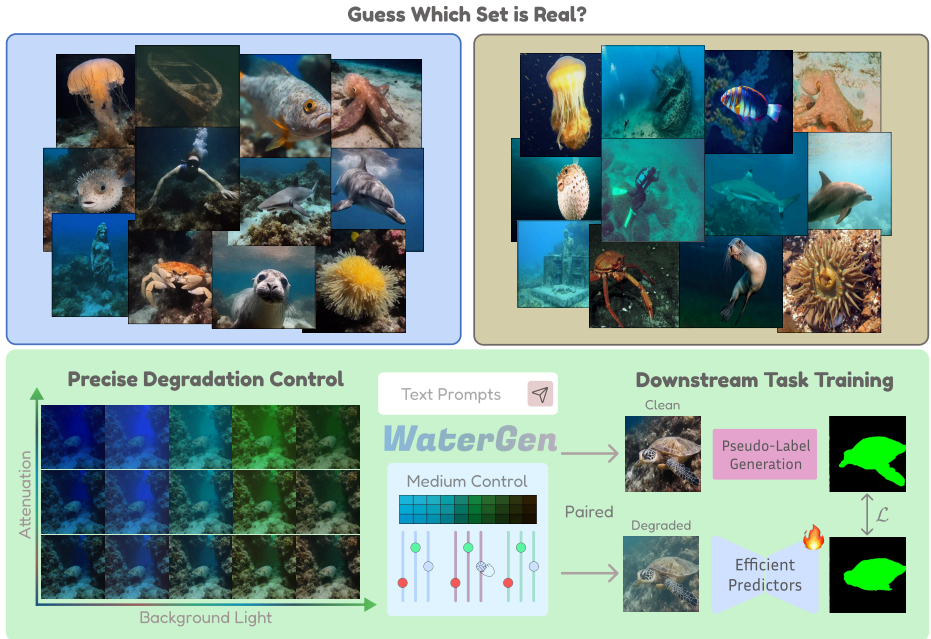}
\caption{\textbf{We introduce WaterGen, an underwater image generation method that enables independent and precise control over medium degradation.} Taking text descriptions and physical water parameters as inputs, WaterGen synthesizes diverse, high-fidelity underwater scenes (\textbf{top-left}) with accurate medium degradation, achieving a striking resemblance to real underwater images (\textbf{top-right}). Our scene-medium decoupled design allows a single latent scene embedding to be rendered across various water types. This capability positions WaterGen as a powerful synthetic data engine for diverse downstream underwater tasks, where perfectly paired degradation-free scenes drastically improve annotation convenience and quality.}
\label{fig:teasor}
\end{figure}

Exploring underwater environments is crucial for marine biology, archaeological preservation, and offshore industrial inspection. However, the efficacy of computer vision algorithms in these domains—such as object detection, image restoration, and segmentation—is severely bottlenecked by the scarcity of high-quality, diverse labeled data. Unlike terrestrial imaging, underwater photography is plagued by complex physical degradation, including wavelength-dependent attenuation, scattering, and low contrast, making large-scale dataset collection both expensive and logistically challenging. To make matters worse, collecting underwater data at scale is challenging, requiring specialized platforms and equipment (e.g. divers, ROVs/AUVs, calibrated lighting) and operating under limited visibility, harsh conditions, and safety/regulatory constraints. Even when imagery is captured, obtaining reliable ground truth, especially for depth~\cite{yu2022udepth,rajyaguru2026polardepth}, geometry~\cite{wu2023low,wu20233d}, and detailed annotations, is costly, slow, and often impractical.

Current approaches to underwater data augmentation and synthesis generally fall into three categories: physics-based rendering, style transfer networks, and image generation models. 
Physics-based methods use the underwater image formation model (UIFM) to degrade in-air images, but because real underwater scenes rarely have scattering-free ground truth, they are typically built on terrestrial datasets, creating a semantic gap and limiting underwater-native scene diversity. 
Data-driven style transfer methods (e.g. GAN-based translation), while popular, typically rely on a fixed scene content source domain. 
Because they are designed to preserve the semantic structure of the input image, they cannot increase the diversity of scene content.
Furthermore, most of these methods neglect the depth-dependent nature of underwater optical physics, reducing the simulation to a superficial color style transfer rather than an accurate medium transition. 
Large-scale generative models, particularly diffusion models, demonstrate impressive capability in synthesizing diverse scene content. However, these models lack explicit knowledge of underwater image formation. Consequently, they often hallucinate results with low physical fidelity, exhibiting a significant domain gap from real underwater environments. 
In addition, text prompts alone do not enable precise control of medium parameters, and the entanglement between scene content and scattering effects prevents independent adjustment of water conditions without altering scene geometry.

\begin{figure}[t]
    \centering
    \includegraphics[width=\linewidth]{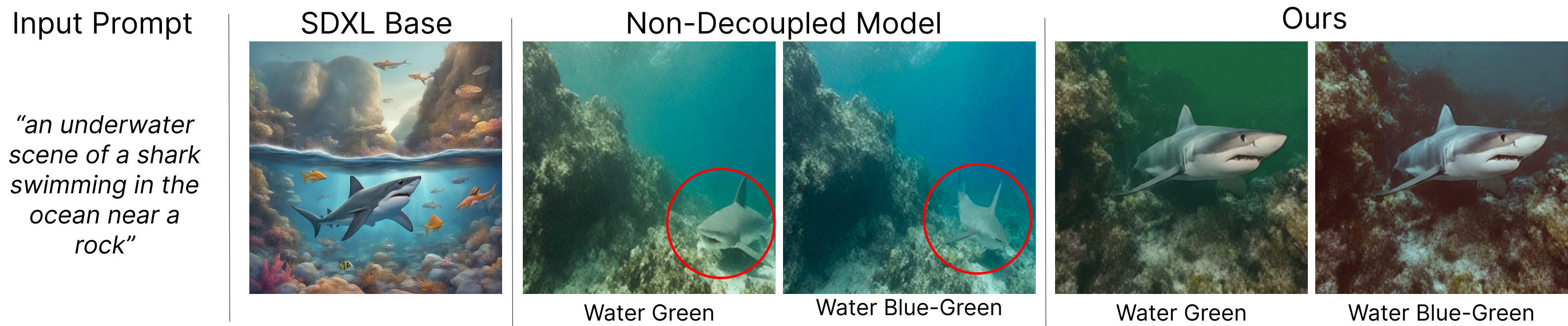}
    \caption{\textbf{Advantage of our WaterGen decoupled generation model.} The base diffusion model (SDXL~\cite{podell2023sdxl}) often produces unrealistic and overly stylized images. A model that is fine-tuned on underwater images without scene and medium decoupling produces inaccurate medium control and can alter object structure when the requested water appearance changes. WaterGen preserves the scene content while changing the medium effects, supporting independent control of scene and water appearance.}
    \label{fig:supp_nondecoupled_comparison}
\end{figure}

To overcome these limitations, we build on pretrained diffusion models and propose WaterGen (Fig.~\ref{fig:teasor}), a scene–medium decoupled framework for underwater image generation that bridges the gap between semantic diversity and physical fidelity. 
Given a natural-language prompt and water parameters (e.g. transmission and backscattering), our method synthesizes high-quality underwater images with realistic, underwater-native structures and physically consistent degradation effects. 
Our key idea is to decouple underwater image synthesis into two controllable factors: scene content and water-medium effects. Scene content generation governs image content and structure and thus relies on  diverse global semantics. 
By contrast, water-medium effects are local in nature and require accurate modeling of pixel-wise geometry and physical parameters.
Based on this view, we decompose generation into two sequential stages. 
First, \textbf{Semantic Scene Latent Diffusion} generates latent representations that capture diverse and plausible underwater scene geometry and layout without imposing medium degradation. 
Second, \textbf{Multi-scale Medium-conditioned Decoding} applies detailed, controllable, and physically grounded attenuation and scattering effects during decoding.
This decoupled design lets each stage be optimized with independent datasets and objective-specific losses: we fine-tune the diffusion U-Net on restored clear underwater images to adapt scene structure, while training spatially aware decoding using physically simulated data to enforce accurate medium control. As shown in~\cref{fig:supp_nondecoupled_comparison}, WaterGen enables controllable and realistic generation that current diffusion-based methods fail to achieve: the base SDXL model often produces unrealistic, stylized, or cartoonish underwater images, while a non-decoupled fine-tuned model provides only weak medium control and can alter object structure when the requested water appearance changes. In contrast, WaterGen preserves the underlying scene content while changing only the medium effects, enabling independent control of scene and water appearance. WaterGen therefore serves as a scalable data engine for producing effectively unlimited amounts of diverse, physically grounded underwater imagery, helping alleviate the data bottleneck in underwater computer vision.

Overall, our main contributions are summarized as follows:
\begin{enumerate}
    \item \textbf{We propose a novel scene-medium decoupled underwater image generation framework.}
    By decoupling semantic scene synthesis from medium degradation and adopting a tailored isolated training strategy, our method preserves the diversity of modern generative models while substantially improving the realism and controllability of underwater medium effects.
    To the best of our knowledge, this work is the first to introduce accurate medium control into diffusion-based underwater image generation.
    \item \textbf{We design a physics-aware decoding mechanism that achieves fine-grained medium control.} Based on the insight that degradation is fundamentally a spatial phenomenon, we inject multi-scale medium conditions exclusively into the latent decoding process. This ensures that the generated attenuation and scattering effects adhere to the underwater image formation model without compromising the semantic integrity of the scene.
    \item \textbf{We show that medium-controlled underwater image generation directly improves downstream vision tasks.} By generating paired water-free ground truth and physically degraded underwater images at scale, our framework provides effective training data for task-specific learning. Experiments demonstrate clear gains in downstream applications, including underwater image restoration and semantic segmentation, validating the utility of accurate and controllable synthetic data generation.
    
\end{enumerate}

\vspace{-5pt}
\section{Background and Related Work}
\label{sec:related_work}

\textbf{Underwater Image Synthesis and Style Transfer.}
Underwater image synthesis aims to simulate or generate realistic underwater images that conform to the optical physics of light propagation in scattering media~\cite{islam2024computer}. 
Recent underwater task works for camouflaged instance segmentation~\cite{wang2026ucis,he2025navmoe,wu2024marvis},  dense prediction~\cite{vit_uwa2026,yuan2025learning,xiong2024event3dgs,cai2021toward}, and salient object detection~\cite{fscdiff2025,wu2025viewactive}, highlight the need for diverse, controllable underwater data synthesis with reliable dense annotations.
Model-based image synthesis~\cite{ueda2019underwater, kaneko2024phiswid, blasinski2016threeparameter,desai2021ruig, desai2024rsuigm, siddique2025aquafuse, lv2025uwstereo} simulates image degradation using underwater imaging equations that capture attenuation and backscatter during light propagation in water. Because these methods are interpretable and controllable, they remain widely used for synthetic data generation. Typically, they take a clean image and a depth map as input and then apply degradation under assumed water parameters. Examples include RUIG~\cite{desai2021ruig},  RSUIGM~\cite{desai2024rsuigm}, and AquaFuse~\cite{siddique2025aquafuse}, all of which extend physical modeling to produce more realistic underwater imagery. However, such approaches depend on prior clean images, offer limited scene diversity, inherit domain bias from terrestrial data, and rely on simplified assumptions about underwater environments.

On the other hand, data-driven methods~\cite{li2017watergan, wang2019uwgan, zhao2021unpaired, zhao2021synthesis, yang2023underwater, xu2023underwater} learn to synthesize underwater imagery from real underwater data examples. FunieGAN~\cite{islam2020fast}, WaterGAN~\cite{li2017watergan}, and image-to-image translation models such as Pix2Pix~\cite{isola2017image} and CycleGAN~\cite{zhu2017unpaired} synthesize underwater appearance by transferring color cast and haze styles from underwater images. These methods increase dataset diversity but often capture only the  visual appearance rather than the underlying physics, leading to physically inconsistent results. Depth-guided approaches such as UStyle~\cite{siddique2025ustyle} improve realism by linking degradation to depth, but medium effects remain entangled and difficult to control. More fundamentally, because style-transfer methods can only transform existing images, they remain limited in scene diversity.

\noindent\textbf{Latent Diffusion Models for Underwater Image Synthesis.} Denoising Diffusion Probabilistic Models (DDPMs)~\cite{ho2020denoising, song2020score} generate data by reversing a noise corruption process. To reduce the high computational cost of pixel-space diffusion~\cite{dhariwal2021diffusion, saharia2022photorealistic}, Latent Diffusion Models (LDMs) such as Stable Diffusion~\cite{rombach2022high} perform diffusion in a compressed latent space encoded by a pre-trained Variational Autoencoder (VAE)~\cite{kingma2013auto}. This latent formulation enables efficient high-quality image synthesis while preserving semantic structure~\cite{ke2025marigold, wang2023breathing,wu2026real2sam2real,wang2025flash,cai2025parametric,jia2024dginstyle}.
Controllable frameworks such as ControlNet~\cite{zhang2023controlnet} and T2I-Adapter~\cite{mou2024t2i} further improve generation by incorporating conditions such as depth, edges, and segmentation, enabling scene-level control by combining semantic prompts with structural guidance.
In the underwater domain, Atlantis~\cite{zhang2024atlantis} extends Stable Diffusion with a Depth2Underwater ControlNet, while TIDE~\cite{lin2025tide} introduces a unified framework for generating underwater images together with dense annotations such as depth, segmentation, and edge maps.
However, these methods remain limited by domain bias and the lack of explicit underwater image formation modeling, often producing stylized \textit{clear water} scenes instead of physically accurate scattering, attenuation, and turbidity, while offering limited control over water optical properties.

\section{Method}
\label{sec:method}

\subsection{Problem Setting: Underwater Image Formation}
The Jaffe-McGlamery (JM) model~\cite{jaffe1990JM} provides a foundational description of underwater image formation by modeling light absorption and scattering in water. 
Subsequently,~\cite{akkaynak2018revised} refined this formulation by distinguishing the attenuation coefficients of direct transmission and backscatter, improving physical accuracy for underwater imaging. A common formulation is:
\begin{equation}
\label{eq:UIFM_Eq}
I_c = J_c \cdot e^{-\beta^D_c d} + B_c^\infty \cdot (1 - e^{-\beta^B_c d}),
\end{equation}
where \( c \) $\in$ \{R, G, B\} represents the color channel; \( I \) represents the image captured underwater by the camera of a scene at distance \(d \); \( J \) is the corresponding clear scene radiance without water; \( B^\infty \)  is the background light or water color at infinity; and the two parameters \( \beta^D \) and \( \beta^B \) represent the attenuation and backscatter coefficients, respectively.

\begin{figure}[t]
  \centering
  \includegraphics[width=0.98\linewidth]{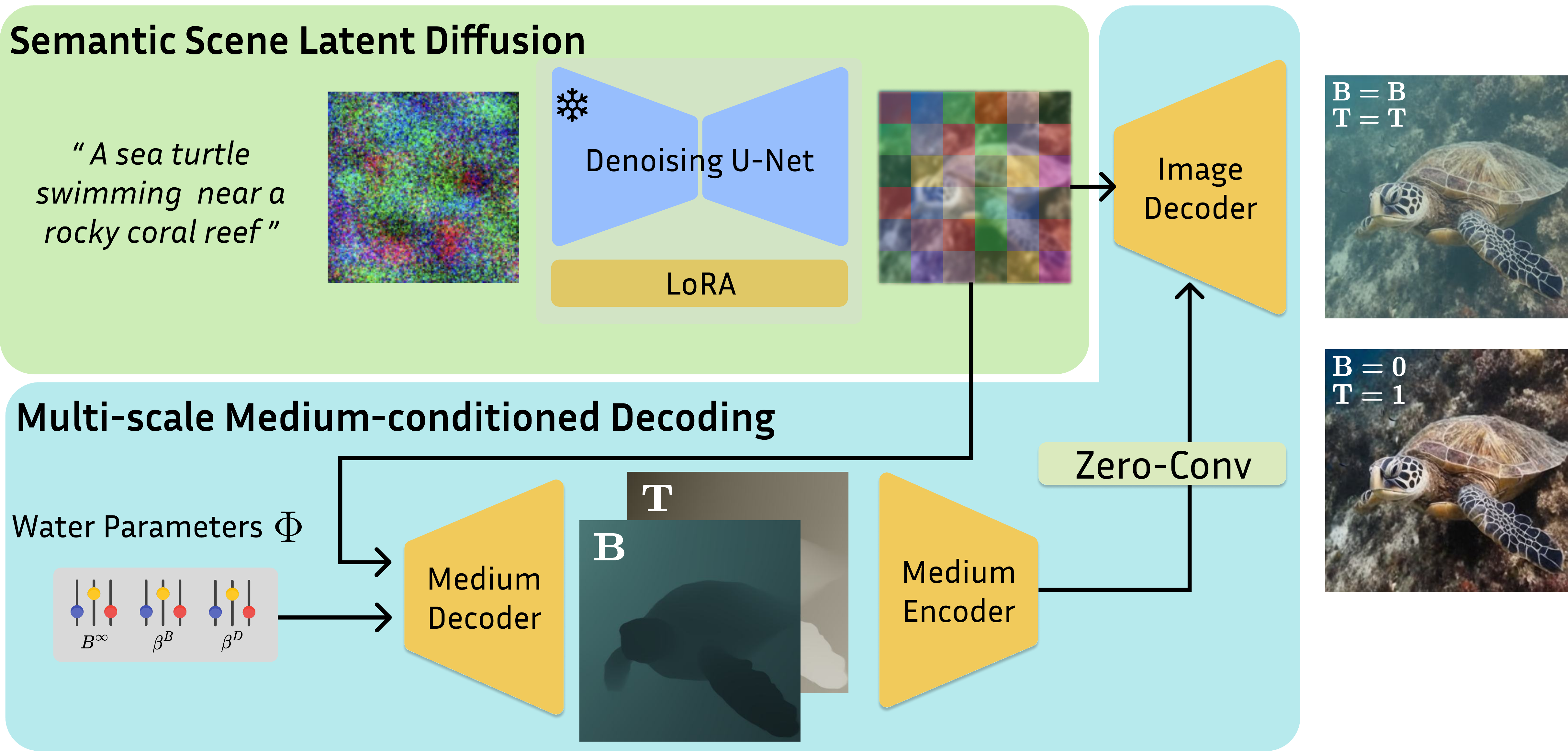}
  \caption{\textbf{Overview of the WaterGen pipeline.} Given a text prompt and water parameters $\Phi$, WaterGen decouples scene generation from medium control to synthesize realistic underwater images. A LoRA-adapted denoising U-Net first produces clean scene latents from text. A medium-conditioned decoder then applies attenuation and scattering consistent with $\Phi$. This design natively supports one scene, multiple waters: the same scene latent can be decoded under different water conditions by varying $\Phi$.
}
  \label{fig:infer_pipeline}
\end{figure}

Drawing from the underwater image formation model in \cref{eq:UIFM_Eq}, two key insights emerge. First, wavelength- and depth-dependent attenuation and scattering are mainly determined by a few intrinsic water parameters ($\Phi=\{B^\infty, \beta^D,  \beta^B\}$). Second, this medium-induced degradation is an extrinsic physical effect on the radiance of the scene, not an intrinsic semantic property of the scene itself.
Therefore, we train a latent diffusion model $\mathcal{G}$ that generates underwater images from a text prompt $P$ while independently controlling the medium with physically meaningful parameters $\Phi$, 
$
I_{gen} = \mathcal{G}\left(  P, \ \ \Phi \right)$,
such that the generated image matches the prompt semantics while exhibiting physically consistent degradation under the specified water conditions.

\subsection{Our Core Idea: Decoupling Scene Generation \& Medium Control}
Our core idea is to decouple scene generation from medium control by treating underwater imagery as the combination of two components: semantic scene content (what is in the image) and participating-medium effects (what the water does to the image).
Existing text-to-image models often entangle these factors into a single learned “underwater style,” which makes medium control weak and biased towards dominant training appearances (e.g. clear blue/green water), even when the prompt specifies challenging turbidity or color conditions.
A natural attempt is to inject medium conditions into the diffusion backbone using established controllable generation mechanisms such as ControlNet~\cite{zhang2023controlnet}, T2I-Adapter~\cite{mou2024t2i}, or by concatenating condition maps with diffusion noise~\cite{ke2025marigold}. These tools are primarily designed for structural guidance (edges, depth, segmentation) and are ill-suited for underwater degradation, which requires radiometrically accurate, spatially varying, pixel-level modulation. 
In our experiments, direct medium injection led to inconsistent medium control and sometimes reduced generation quality.

To resolve this mismatch, we adopt a principled separation within the latent diffusion framework: semantic scene synthesis is performed in the latent diffusion backbone, while medium modulation is executed in the decoding module.
This design follows the distinct nature of the two tasks. 
Diffusion is the right place for scene synthesis because it is a semantic generative task requiring diversity, compositional reasoning, global context, and strong prompt alignment. 
In contrast, medium simulation is best treated as a reconstruction-time radiometric process requiring per-pixel precision and direct supervision from physics. Consequently, we keep the diffusion stage responsible for producing a clean, water-free semantic latent image that preserves the diversity and text conditioning, and we relocate physical injection to the decoder, where spatial fidelity is crucial and physically meaningful maps derived from input water parameters can modulate pixel reconstruction. 
This decoupled design enables precise, independent control over water conditions, enforces physics-consistent water degradation, and preserves semantic quality by preventing the semantic manifold of the diffusion latent space from being distorted by learning optically degraded domain statistics.
Our decoupled framework naturally supports "one scene, many waters": the same semantic latent can be decoded under different water conditions by varying the input parameters. This enables efficient generation of perfectly aligned clean/degraded pairs, providing scalable paired data for downstream tasks while keeping scene content fixed.

\begin{figure}[t]
  \centering
  \includegraphics[width=0.98\linewidth]{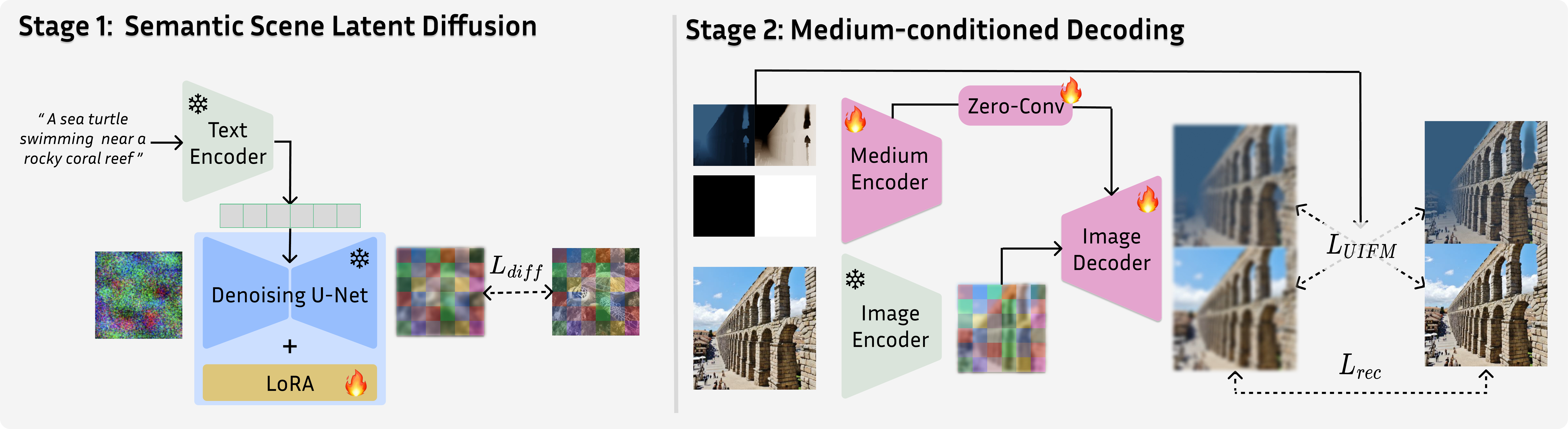}
  \caption{\textbf{Training pipeline of WaterGen.} We adopt a two-stage isolated training strategy to decouple scene generation from medium degradation. Stage 1 fine-tunes a latent diffusion backbone with LoRA on restored, water-free underwater images to learn scene geometry and layout. Stage 2  independently trains the decoder on physically accurate degraded terrestrial data to learn to conditionally inject water medium effects into the clean image latent. 
}
  \label{fig:train_pipeline}
\end{figure}

\vspace{-10pt}
\subsection{Pipeline Architecture}
Our proposed computational pipeline is outlined in Fig. \ref{fig:infer_pipeline}; it addresses physics-informed underwater image generation with precise medium control by strictly decoupling semantic scene generation from physical medium degradation simulation. 
Given a text prompt and a set of water parameters, the model first performs \textbf{Semantic Scene Latent Diffusion} to generate a clean latent representation of the underlying scene, without underwater effects. This stage is built on an SDXL backbone~\cite{podell2023sdxl} fine-tuned to model \textit{pristine, water-free} underwater scene content, allowing it to focus on semantic structure and diversity rather than medium degradation.
The clean scene latent is then passed to our proposed \textbf{Multi-scale Medium-Conditioned Decoder}. 
Conditioned on the input water parameters $\Phi$ and the clean image latent, a medium decoder, which contains a standard diffusion decoder and depth estimation module~\cite{bochkovskii2024depthpro}, produces the corresponding backscattering and transmission maps, denoted by ($B, T$).
These physically grounded pixel-level medium descriptors are subsequently encoded and introduced into the image decoder as explicit spatial conditions for reconstructing the final image.
Specifically, based on the definitions in the Underwater Image Formation Model (\cref{eq:UIFM_Eq}), the decoder uses the input water parameters $\Phi$ to formulate pixel-level medium descriptions—namely the backscattering map and transmission map. These physically derived maps serve as explicit spatial conditions that modulate the decoding of the clean latent. 
The final decoded output preserves the semantic structure of the generated scene while accurately matching the specified water conditions.

\subsection{Two-Stage Isolated Training Process}
We now describe the isolated two-stage training process illustrated in Fig. \ref{fig:train_pipeline}.

\subsubsection{Tailored Data Curation and Synthesis} The isolated training strategy empowers us to employ  optimal data sources tailored to the different training objectives of the diffusion and decoding components. For fine-tuning the semantic scene diffusion stage, we aggregate a large-scale collection of pristine underwater scene representations by applying the state-of-the-art restoration model, SLURPP~\cite{wu2025single}, to multiple publicly available real-world underwater datasets. The corresponding textual captions are generated using the BLIP~\cite{li2023blip} vision-language model. Conversely, for the medium-conditioned decoding stage, the isolation from the semantic backbone grants us the flexibility to leverage large-scale terrestrial datasets, which offer the "water-free" ground truth to synthesize precise underwater optical degradation. To ensure physical precision, we synthesize diverse and realistic degradation effects by projecting these terrestrial scenes through the underwater image formation model, utilizing water parameters randomly sampled from empirical distributions derived from extensive real-world measurements (following SLURPP~\cite{wu2025single}). Crucially, this synthetic generation pipeline also facilitates the implementation of the stochastic medium noise injection mechanism detailed below.

\vspace{-10pt}
\subsubsection{Stage 1: Clean-Target Diffusion Fine-Tuning}
The primary objective of training {\textbf{Semantic Scene Latent Diffusion}} is to adapt the diffusion backbone to generate clear and degradation-free underwater scene semantics conditioned on text prompts, without compromising the generative quality of the original pretrained model.
To achieve this, we construct a large-scale training dataset by applying a state-of-the-art underwater image decoupling pipeline to extensive publicly available real-world underwater datasets. This process yields a vast collection of pristine, medium-free underwater scene images, paired with descriptive captions generated via an image captioning pipeline.

During training, we freeze the VAE and original U-Net weights and fine-tune only the LoRA layers inserted into the attention blocks. We then optimize the model with the standard diffusion denoising objective, using the \emph{clean} scene images as training targets. 
By calculating the gradients solely against the clean scene latents rather than the degraded underwater images, we establish a strong inductive bias. This forces the diffusion model to internalize ``water-free'' underwater semantics (e.g. coral structures, divers) while implicitly treating any potential residual medium artifacts as noise to be removed. This strategy effectively preserves the high-frequency spatial fidelity of the latent space, preventing the latent space degradation typically associated with fine-tuning on optically degraded domain data.

\subsubsection{Stage 2: Noise-Injected Medium-Conditioned Decoding}
The training of the \textbf{Multi-scale Medium-Conditioned Decoder} is isolated from the semantic diffusion backbone. Based on the insight that medium simulation is orthogonal to semantic generation, we treat the decoding process as a standalone pixel-space reconstruction task. This isolation gives us the flexibility to utilize large-scale terrestrial datasets—which are free from underwater domain biases—to rigorously train the physical simulation capabilities of the decoder.

We construct a specific training pipeline using high-quality terrestrial images as ground truth scenes $J$. To ensure physical diversity, we project these scenes through the underwater image formation model using water parameters $\Phi=\{B^\infty, \beta^D, \beta^B\}$ sampled from a database of real-world measurements. This gives us precise training triplets $(J, I, \Phi)$ with substantial variability, providing the decoder with exact physical supervision.

To ensure that the decoder is robust to the unavoidable residual medium artifacts present in the Stage 1 clean latents, we employ a \textbf{Stochastic Medium Noise Injection} strategy. During training, we apply random and slight physical degradations to the input images before encoding. This produces "perturbed" latents that mimic the imperfect output of the diffusion model. This mechanism forces the decoder to disregard the implicit noise or "style" in the latent embedding and condition the reconstruction strictly on the injected physical parameters $\Phi$, ensuring that the model learns the underlying physical laws rather than a simple identity mapping.

We apply a dual-forward pass mechanism for each latent $z$ in every training iteration. 
For the first pass, the medium conditions are set to a non-degraded state ($\Phi=\mathbf{0}$). The decoder must reconstruct the water-free image $\hat{J}$.
For the second pass, the medium conditions are set to the sampled degradation parameters $\Phi_{input}$. The decoder must synthesize the corresponding physically degraded image $\hat{I}$. To rigorously enforce the physical validity of the synthesis, we introduce a Bidirectional Physics-Consistency Constraint. We not only minimize the reconstruction error of the forward simulation ($\hat{J} \rightarrow I_{gt}$) but also enforce an inverse consistency constraint ($\hat{I} \rightarrow J_{gt}$).
\begin{equation} 
\mathcal{L}_{UIFM} = \left| \left( \hat{J} \cdot T + B \right) - I_{gt} \right|_1 +  \left| \frac{\hat{I} - B}{T} - J_{gt} \right|_1
\end{equation}
where $T = e^{-\beta^D_c d}$, $B = B_c^\infty \cdot (1 - e^{-\beta^B_c d})$.
The total objective is formulated as:
\begin{equation}
\mathcal{L}_{stage2} = \mathcal{L}_{rec}(I) + \mathcal{L}_{rec}(J) + \lambda_{uifm}\mathcal{L}_{UIFM}
\end{equation}
where the reconstruction term $\mathcal{L}_{rec}$ is defined as a weighted sum: $\mathcal{L}_{rec} = \lambda_{1}\mathcal{L}_{1} + \lambda_{ssim}\mathcal{L}_{SSIM} + \lambda_{lpips}\mathcal{L}_{LPIPS}$. In our implementation, we set $\lambda_{1}=1.0$, $\lambda_{ssim}=1.0$, $\lambda_{lpips}=0.5$, and  $\lambda_{uifm}=0.3$.

\begin{figure}[t]
\centering
\includegraphics[width=\linewidth]{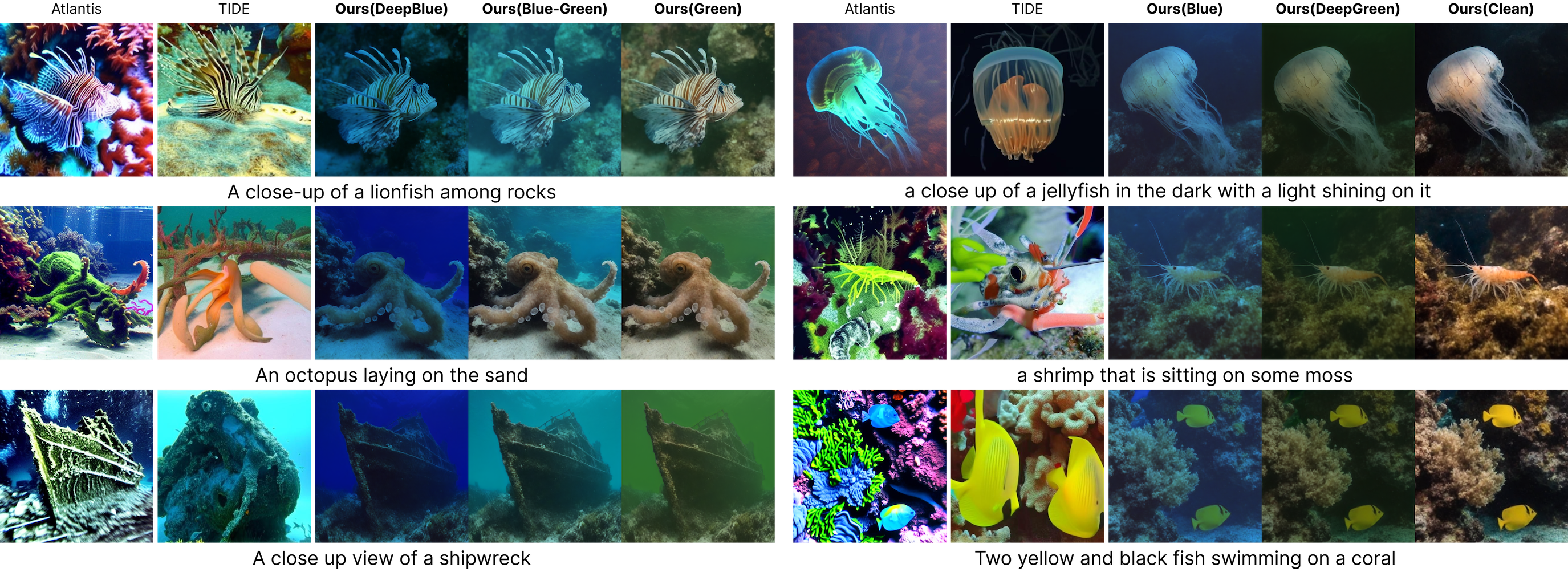}
\caption{\textbf{Qualitative results on synthesis fidelity and diverse water types compared to Atlantis~\cite{zhang2024atlantis} and TIDE~\cite{lin2025tide}.} We visualize our pipeline's outputs across six common water types (DeepBlue, Blue, Blue-Green, Green, DeepGreen, and Clean) from the UF7D dataset~\cite{siddique2025ustyle}. Because the baselines lack precise medium-degradation control, we control water medium using text prompts. For fairness, we provide Atlantis with depth maps obtained from~\cite{bochkovskii2024depthpro} using our generated images.
}
\label{fig:main_qual}
\end{figure}

\section{Experiment Results}
\label{sec:results}

\subsection{Datasets and Experimental Setup}
We initialize our latent diffusion model using SDXL (Base)~\cite{podell2023sdxl}, while both the medium encoder and the image encoder-decoder are initialized with the pre-trained SDXL VAE weights. Low-Rank Adaptation (LoRA)~\cite{hu2022lora} is applied to the diffusion U-Net, configuring the adapters with a rank of 32 and an alpha of 16. For Stage 1 training, we introduce WaterGen-Clean, 
which contains text-captioned, degradation-free underwater images. Raw images sourced from six standard training sets (UIIS10K~\cite{UIIS10K_Dataset_2025}, USIS10K~\cite{lian2024USIS}, SUIM~\cite{islam2020semantic}, USOD10K~\cite{hong2023usod10k}, UIEB~\cite{li2019UIEB}, and UF7D~\cite{siddique2025ustyle}) are restored using the state-of-the-art SLURPP~\cite{wu2025single} model and captioned via pre-trained BLIP-2~\cite{li2023blip} to align with TIDE~\cite{lin2025tide} and Atlantis~\cite{zhang2024atlantis} baselines. In particular, while the SLURPP restorations are not universally devoid of degradation artifacts, the model consistently yields high-quality outputs. 
For Stage 2 training, we synthesize underwater degradation on high-quality terrestrial images via an Underwater Image Formation Model ( ~\cref{eq:UIFM_Eq}). The resulting $(J, I, \Phi)$ triplets provide the medium-conditioned decoder with precisely annotated physical medium data. The Stage 1 diffusion model and the Stage 2 conditional latent decoder are trained sequentially on a single NVIDIA A6000 GPU. Both stages employ an identical learning rate of $10^{-5}$, and each training stage takes approximately 1 day to converge.

\subsection{Underwater Image Generation}

We quantitatively and qualitatively compare our method with two state-of-the-art underwater image generation models, Atlantis~\cite{zhang2024atlantis} and TIDE~\cite{lin2025tide}, utilizing 5,451 underwater scene captions from the SynTIDE~\cite{lin2025tide} dataset. The CLIP Score~\cite{radford2021clip} is used to evaluate the semantic consistency of the generated images. Because synthesized underwater degradation intrinsically lowers standard visual quality scores, we isolate this confounding factor to fairly assess the visual fidelity of the underlying scene generation. Specifically, we compare our degradation-free outputs ($B=0, T=1$) with baseline generations post-processed by the state-of-the-art SLURPP restoration model, using two standard reference-free metrics: UIQM~\cite{panetta2015UIQM} and MUSIQ~\cite{ke2021musiq}. Specifically, UIQM focuses on evaluating underwater-specific attributes such as colorfulness, sharpness, and contrast, whereas MUSIQ provides a comprehensive, deep-learning-based assessment of overall multi-scale perceptual quality.

As shown in ~\cref{tab:generation_metrics}, compared to baselines—which suffer from decreased semantic adherence after fine-tuning on heavily degraded underwater images—our WaterGen achieves significantly higher visual fidelity in native clear scene generation while maintaining robust semantic consistency. More importantly, owing to our scene-medium decoupled design, we not only prevent underwater degradation effects from corrupting the latent space during fine-tuning but also enable the injection of pixel-level medium conditions during the decoding of latent embeddings. This facilitates physically accurate, degradation-controlled generation and re-editing.

As illustrated in Fig.~\ref{fig:main_qual}, in contrast to the baselines, our pipeline can decode a single latent diffusion embedding under varying medium degradation conditions, thereby yielding highly realistic underwater degradations across a diverse array of water types. 
TIDE~\cite{lin2025tide} can produce multimodal outputs (underwater images, depth maps, and masks), but it often fails to generate coherent scene and object structures.
Meanwhile, Atlantis leverages text prompts for semantic control and depth maps for scene structure. Although it preserves the scene structure relatively well, the generated textures exhibit overly vibrant and stylized artifacts.
Because both methods are trained on degraded underwater images without decoupling water effects, the visual quality and text-image alignment of their outputs are suboptimal, as shown in~\cref{tab:generation_metrics}.
We show additional analysis on generation diversity, medium control, and non-decoupled baseline comparisons in the supplementary.

\begin{table}[t]
\centering
\caption{Quantitative comparison against Atlantis~\cite{zhang2024atlantis} and TIDE~\cite{lin2025tide} on clear underwater scene generation. Visual quality is evaluated using UIQM~\cite{panetta2015UIQM} and MUSIQ~\cite{ke2021musiq}, and text-image alignment is evaluated using CLIP~\cite{radford2021clip}. To isolate scene visual fidelity from underwater degradation, we compare our degradation-free generations ($B=0, T=1$) against baseline outputs that have been post-processed by a state-of-the-art underwater image restoration model SLURPP~\cite{wu2025single}. The numbers presented in the table denote the mean and standard deviation (mean $\pm$ std) calculated across five distinct random seeds.}
\label{tab:generation_metrics}
\resizebox{0.90\linewidth}{!}{
\begin{tabular}{@{}lcccc@{}}
\toprule
{Method} & {UIQM} $\uparrow$ & {MUSIQ} $\uparrow$ & {CLIP Score} $\uparrow$ & {Controllability} \\ \midrule
Atlantis~\cite{zhang2024atlantis}    & $2.8338 \pm 0.1927$                   & $67.5437 \pm 1.7373$                    & $0.2457 \pm 0.0274$                          & Text-only               \\
TIDE~\cite{lin2025tide}           & $2.3725 \pm 0.3816$                  &  $66.4304 \pm 2.1780$                     & $0.2305 \pm 0.0118$             & Text-only               \\ \midrule
\textbf{Ours}   & $\mathbf{3.0239 \pm 0.1317}$          & $\mathbf{69.2638 \pm 0.8813}$          & $\mathbf{0.2614 \pm 0.0073}$    & \textbf{Text + Medium}  \\ \bottomrule
\end{tabular}
}
\end{table}

\begin{table}[t]
\centering
\caption{{Quantitative comparison on UIIS10K~\cite{UIIS10K_Dataset_2025} dataset using UIQM~\cite{panetta2015UIQM} and MUSIQ~\cite{ke2021musiq} reference-free metrics.} Higher is better ($\uparrow$).}
\renewcommand{\arraystretch}{1.2}
\setlength{\tabcolsep}{6pt}
\resizebox{0.90\linewidth}{!}
{\begin{tabular}{lcccc}
\toprule
{Metric} & {WaterNet~\cite{li2019UIEB}} & {Phaseformer~\cite{khan2024phaseformer}} & {DeepWaveNet~\cite{sharma2021deepwavenet}} & {Histoformer~\cite{peng2024histoformer}} \\
\midrule
UIQM Baseline & 3.067 & 2.239 & 2.685 & 2.994 \\
UIQM Ours & \textbf{3.141} & \textbf{2.272} & \textbf{2.731} & \textbf{3.048} \\
\midrule
MUSIQ Baseline & 66.866 & 67.400 & 66.843 & 66.178 \\
MUSIQ Ours & \textbf{68.322} & \textbf{69.165} & \textbf{67.937} & \textbf{66.924} \\
\bottomrule
\end{tabular}}
\label{tab:uiis10k_uiqm_musiq}
\end{table}

\vspace{-10pt}

\subsection{Downstream Applications}
\subsubsection{Underwater Image Restoration}
To validate the practicality of WaterGen beyond data synthesis, we further evaluate its impact on underwater image restoration. Using our scene–medium decoupled framework, we generate over 20,000 paired underwater images by varying physically meaningful water parameters while preserving scene semantics. This synthetic data is then used as additional training augmentation, exposing restoration models to a broader range of attenuation and backscattering conditions than typically available in real datasets.
As we show in ~\cref{tab:uiis10k_uiqm_musiq}, quantitative evaluation on a real-world benchmark shows that training with our generated data consistently improves restoration performance across multiple representative baselines under UIQM~\cite{panetta2015UIQM} and MUSIQ~\cite{ke2021musiq} metrics. We show more restoration visualizations in the supplementary.
\vspace{-10pt}
\subsubsection{Degradation-Robust Underwater Image Segmentation}

\begin{figure}[t]
    \centering
    \includegraphics[width=\linewidth]{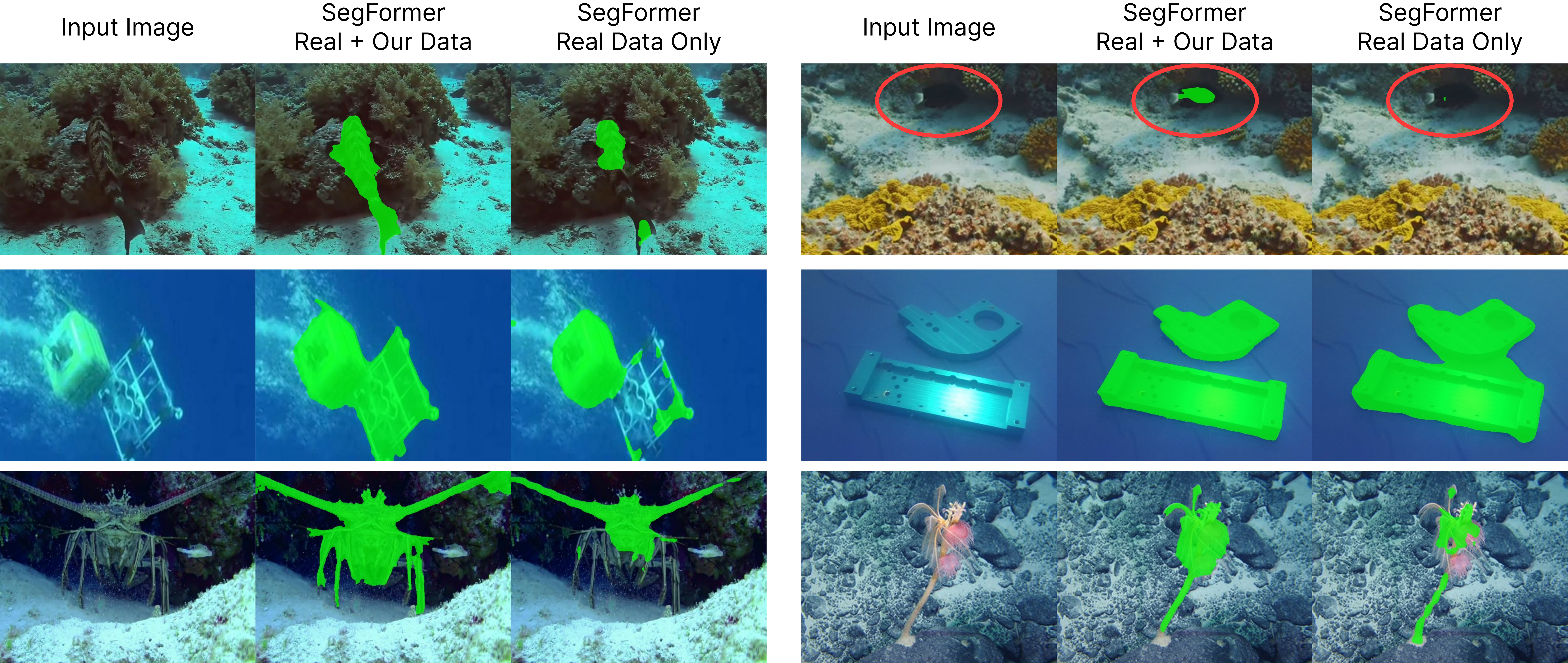}
    \caption{\textbf{Qualitative comparison under strong  underwater degradation.} From left to right: input image, SegFormer~\cite{xie2021segformer} trained on \textit{Real + Our Data}, and SegFormer trained on \textit{Real Data Only}. Adding our data produces cleaner and more complete segmentation masks with sharper boundaries and fewer missed detections, particularly under challenging conditions such as strong turbidity, color shift, and backscattering.}
    \label{fig:qualseg}
\vspace{-10pt}

\end{figure}

We further evaluate our pipeline for underwater semantic segmentation. 
We generate $40,000$ paired clean and degraded underwater images using underwater scene prompts and randomly sampled water medium parameters. We then apply off-the-shelf underwater segmentation models~\cite{lian2024USIS, UIIS_Dataset_2023} to the \textbf{clean images} to extract  pseudo-masks of object classes following TIDE~\cite{lin2025tide} (Fish, Reefs, Aquatic Plants, Wrecks,
Human Divers, and Robots). 
These masks serve as reliable ground-truth supervision for training the models on their corresponding degraded counterparts, enabling efficient construction of a synthetic dataset with accurate annotations in diverse and challenging underwater conditions.
As shown in Tab.~\ref{tab:underwater_seg}, training with our synthetic data consistently improves segmentation performance across architectures and benchmarks. Compared to models trained exclusively on real data, adding our synthetic samples yields clear  gains on both UIIS and USIS10K test datasets.
Qualitative results, shown in Fig.~\ref{fig:qualseg}, further show cleaner, more complete masks, sharper boundaries, and fewer missed detections under severe degradation. 
These results show the effectiveness of our pipeline in generating high-quality supervision and improving segmentation robustness. We show more segmentation visualizations in the supplementary.

\begin{table}[t]

\centering
\caption{Quantitative results of underwater semantic segmentation. We calculate the mean Intersection over Union (mIoU) over six categories (Fish, Reefs, Aquatic Plants, Wrecks, Human Divers, and Robots) following TIDE~\cite{lin2025tide}. Higher is better ($\uparrow$).}
\label{tab:underwater_seg}
\setlength{\tabcolsep}{4pt}
\resizebox{0.90\linewidth}{!}{
\begin{tabular}{llccc}
\toprule
Dataset & Setting & SegFormer (MiT-B4) & Mask2Former (Swin-B) & ViT-Adapter (B) \\
\midrule
\multirow{3}{*}{UIIS~\cite{UIIS_Dataset_2023}}
& Real only    & 70.2 & 72.7 & 73.5 \\
& Real+SynTIDE~\cite{lin2025tide} & 75.4{\textcolor{magenta}{(+5.2)}} & 74.3{\textcolor{magenta}{(+1.6)}} & 75.1{\textcolor{magenta}{(+1.6)}} \\
& Real+Ours    & 75.6{\textcolor{magenta}{(+5.4)}} & 74.0{\textcolor{magenta}{(+1.3)}} & 75.3{\textcolor{magenta}{(+1.8)}} \\
\midrule
\multirow{3}{*}{USIS10K~\cite{lian2024USIS}}
& Real only    & 74.6 & 76.1 & 74.6 \\
& Real+SynTIDE~\cite{lin2025tide} & 76.1{\textcolor{magenta}{(+1.5)}} & 77.1{\textcolor{magenta}{(+1.0)}} & 76.7{\textcolor{magenta}{(+2.1)}} \\
& Real+Ours    & 76.7{\textcolor{magenta}{(+2.1)}} & 76.9{\textcolor{magenta}{(+0.9)}} & 76.9{\textcolor{magenta}{(+2.3)}} \\
\bottomrule
\end{tabular}}
\end{table}

\vspace{-10pt}
\subsection{Ablation Study}

\subsubsection{Medium Injection Mechanism} 
\begin{figure}[t]
\centering
\includegraphics[width=\linewidth]{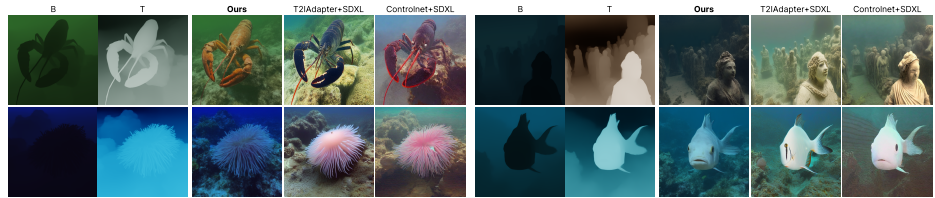}
\caption{\textbf{Qualitative ablation study on different medium injection mechanisms.} While capable of dictating high-frequency structures, ControlNet~\cite{zhang2023controlnet} and T2I-Adapter~\cite{mou2024t2i} fail to accurately control low-frequency scattering color and attenuation.}
\label{fig:ablation_qual}
\end{figure}

Obtaining accurate medium degradation control depends critically on where and how medium conditions are injected.
We therefore compare our method with two widely used conditioning architectures, ControlNet~\cite{zhang2023controlnet} and T2I-Adapter~\cite{mou2024t2i}, using 5,451 SynTIDE~\cite{lin2025tide} captions and five randomly sampled medium conditions per caption. 
We evaluate medium control accuracy by extracting the background light intensity~\cite{song2018rapid} from the generated image and compare it with the input condition.
As shown in Fig.~\ref{fig:ablation_qual} and Tab.~\ref{tab:ablation_bg_light}, both baselines do not accurately control scattering color and attenuation intensity.
Unlike our framework, they do not decouple semantic generation from medium degradation but instead entangle both within the diffusion process. As a result, the medium conditions cannot be enforced faithfully, even when the scene structure is preserved.
Moreover, modeling the effects of water degradation in the denoising U-Net conflicts with the pre-trained “denoise-to-clear” prior, distorting the latent distribution and weakening the overall generative quality.
\vspace{-10pt}

\vspace{-5pt }

\begin{table}[t]
\centering
\caption{Quantitative ablation study on medium injection mechanisms and internal components. To systematically evaluate the accuracy of medium control, we compare our full pipeline against standard adapter-based condition injection mechanisms and ablated training design choices, validating both our architecture and component choices. We measure the Root Mean Square Error (RMSE), Mean Angular Error (MAE), and the CIEDE2000 color difference ($\Delta E_{00} $).}
\label{tab:ablation_bg_light}
\begin{tabular}{lccc}
\toprule
Medium Injection Method & RMSE ($\downarrow$) & MAE ($\downarrow$) & $\Delta E_{00}$ ($\downarrow$) \\
\midrule
ControlNet~\cite{zhang2023controlnet}+SDXL~\cite{podell2023sdxl} & 0.45 & 21.05° & 43.70 \\
T2I-Adapter~\cite{mou2024t2i}+SDXL~\cite{podell2023sdxl} & 0.30 & 9.96° & 30.10 \\
\midrule
Ours (w/o Degradation inject) & 0.07 & 7.01° & 7.27 \\
Ours (w/o Bidir UIFM Consistency) & 0.07 & 7.31° & 6.90 \\
Ours (Full) & \textbf{0.06} & \textbf{4.42°} & \textbf{5.44} \\
\bottomrule
\end{tabular}
\end{table}

\subsubsection{Bidirectional UIFM Consistency and Stochastic Medium Noise Injection} 
As demonstrated in ~\cref{tab:ablation_bg_light}, our random medium degradation injection and bidirectional UIFM consistency self-supervised loss jointly compel the model to ignore the influence of any residual medium information within the latent space during decoding. This prevents the conditional decoder from degenerating into a naive weighted superposition of medium maps, which would otherwise lead to error accumulation. Furthermore, the bidirectional UIFM consistency loss, coupled with a clear-degraded dual forward pass during each training iteration, ensures that the decoder learns a consistent, physically grounded degradation process rather than merely overfitting the synthetic data distribution.
More details on the ablation experiments can be found in the supplementary.
\vspace{-5pt}
\section{Conclusion}
\vspace{-5pt}
\label{sec:conclusion}
We introduced WaterGen, a physically grounded underwater image generation framework that decouples scene semantics from water-medium effects within a latent diffusion pipeline. 
Our decoupled two-stage design combines clean underwater scene latent diffusion with medium-conditioned decoding, enabling precise control over attenuation and backscattering without sacrificing semantic fidelity or scene diversity. 
Beyond realistic and controllable image synthesis, WaterGen serves as a scalable data engine for producing paired underwater data with accurate medium variation.
Extensive experiments demonstrate that our method outperforms existing underwater generative baselines in fidelity and controllability, and that the resulting synthetic data provides consistent benefits for downstream restoration and segmentation tasks. 

\section*{Acknowledgments}
J.W. and Y.A. were supported in part by USDA NIFA sustainable agriculture system program under award no.~20206801231805.
T.W. and C.A.M. were supported in part by the UMD AIM Seed Grant Program, NSF CAREER grant no.~2339616, and ONR grant no.~N00014-23-1-2752. 
M.J.I. was supported in part by the NSF grant no.~2330416.

\clearpage  

%
%
\bibliographystyle{splncs04}
\bibliography{main}
\clearpage
{
  \centering
  \normalfont\normalsize\vskip0.2em{\Large \textbf{Supplementary Material for: ``WaterGen: Decoupling Scene and Medium in Underwater Image Generation'' }\par}\vskip1.0em\par
}

\section{More Visualizations of Generation Diversity}
\begin{figure}[t]
    \centering
    \includegraphics[width=\linewidth]{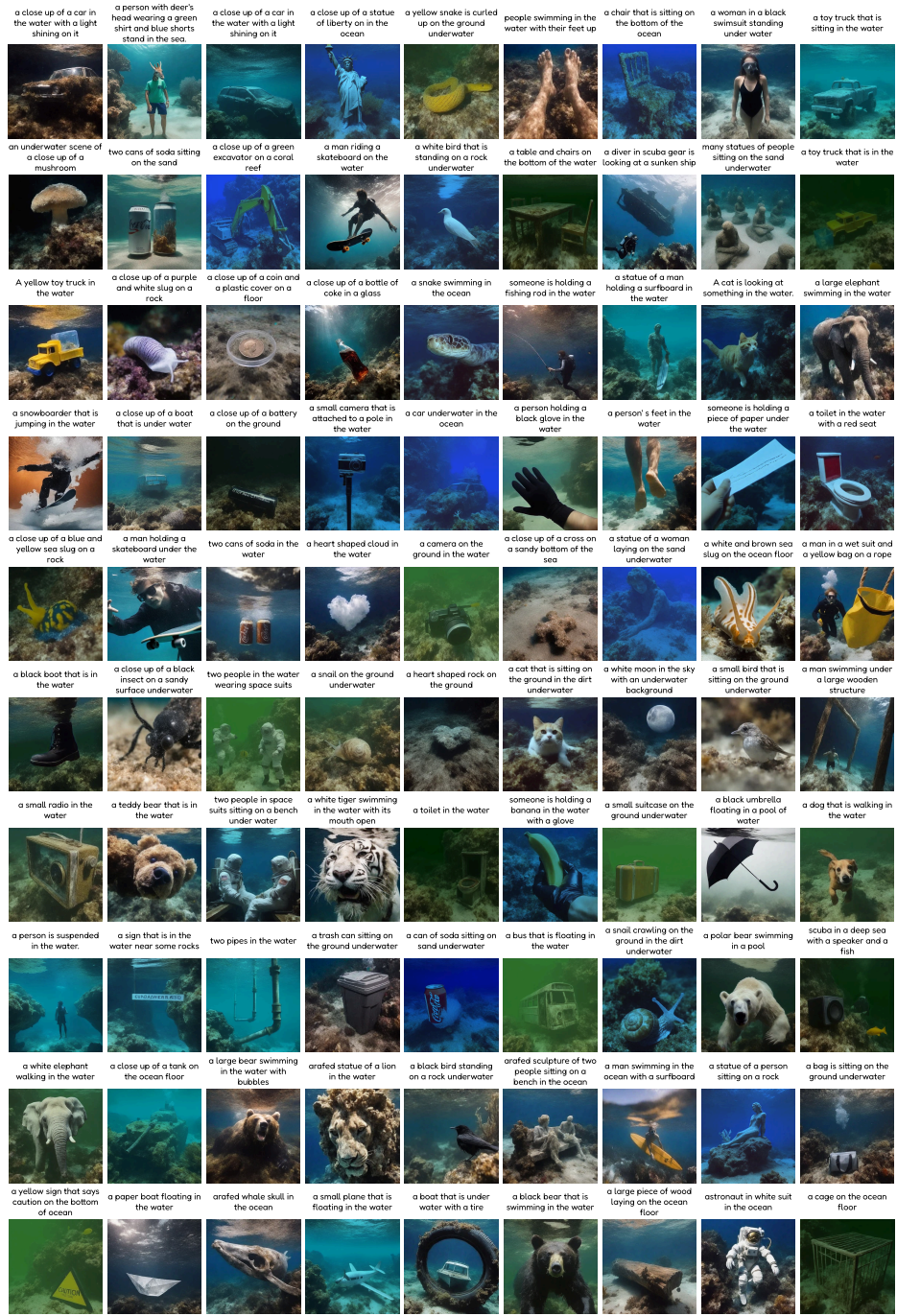}
    \vspace{-30pt}
    \caption{\textbf{Diverse zero-shot generation of underwater environments.} WaterGen demonstrates strong generalization in synthesizing diverse, high-fidelity underwater scenes featuring concepts unseen in the training data.}
    \label{fig:diversity_extra}
\end{figure}

In \cref{fig:diversity_extra}, we present a batch of zero-shot generation results produced by WaterGen on concepts not seen in our training set. To rigorously evaluate the generative diversity and out-of-domain generalization capabilities of our method, we employed a generative large language model (Gemini 3.1 Pro) to automatically generate descriptions of rare or surreal scenes. These prompts—such as "astronaut in white suit in the ocean" and "a small plane that is floating in the water"—served as text conditions alongside the randomly sampled medium parameters. By successfully decoupling scene generation from medium degradation synthesis within a latent diffusion model, WaterGen preserves the generative capability of underlying diffusion model and consistently achieves diverse, high-fidelity underwater image generation, even for highly unconventional scene content.

\section{Non-Decoupled Baseline Results}
To test the importance of decoupling, we train a non-decoupled, parameter-conditioned ablation model on real underwater image-text pairs. We estimate scattering and illumination parameters using ULAP~\cite{song2018rapid} and append them as RGB values to the text prompt. As shown in \cref{tab:supp_nondecoupled_metrics,fig:supp_nondecoupled_comparison}, compared to WaterGen, this baseline performs worse both in image quality and illumination-control accuracy. Additionally, text conditioning cannot reliably preserve object structure when changing the water medium. These results validate the need for our decoupled design.

\begin{table}[t]
\centering
\caption{\textbf{Comparison with a non-decoupled, parameter-conditioned baseline.} Image quality is measured on degradation-free generations, while illumination-control accuracy measures agreement between the requested and generated medium appearance. WaterGen achieves better visual quality and substantially more accurate medium control.}
\label{tab:supp_nondecoupled_metrics}
\resizebox{0.9\linewidth}{!}{
\begin{tabular}{@{}lccc@{}}
\toprule
{Image Quality} & {UIQM} $\uparrow$ & {MUSIQ} $\uparrow$ & {CLIP Score} $\uparrow$ \\ \midrule
Non-Decoupled Model    & $2.68 \pm 0.32$                   & $63.66 \pm 2.16$                    & $0.25 \pm 0.01$                                   \\
\textbf{Ours}   & $\mathbf{3.02 \pm 0.13}$          & $\mathbf{69.26 \pm 0.88}$          & $\mathbf{0.26 \pm 0.01}$     \\ \bottomrule
{Illumination Control Accuracy} & {RMSE} $\downarrow$ & {MAE} $\downarrow$ & $\Delta E_{00}$ $\downarrow$ \\ \midrule
Non-Decoupled Model    & $0.40$  & 35.57°                 & $41.12$                                   \\
\textbf{Ours}   & $\mathbf{0.06}$          & \textbf{4.42°}          & $\mathbf{5.44}$     \\ \bottomrule
\end{tabular}
}
\end{table}

\section{Decoder Ablation on Degradation-Free Image Generation}
\begin{figure}[h]
    \centering
    \includegraphics[width=\linewidth]{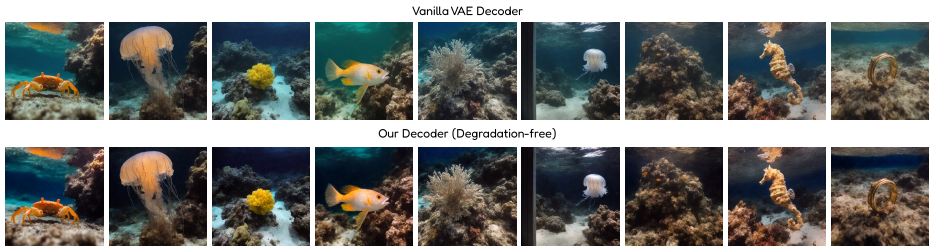}
    \caption{\textbf{Qualitative decoder ablation on degradation-free image generation.} All images are generated by decoding the exact same scene latents produced during WaterGen's latent diffusion stage. \textbf{Top row:} Results from the vanilla VAE decoder. \textbf{Bottom row:} Results from our medium-conditional decoder, conditioned on strictly degradation-free parameters ($B^\infty=0, \beta=0$). Notably, the images decoded by the vanilla VAE still exhibit residual water effects and color shifts. In contrast, our conditional decoder accurately synthesizes precise, completely clean scenes, demonstrating the effectiveness of our decoupling strategy.}
    \label{fig:decoder_ablation}
\end{figure}

In \cref{fig:decoder_ablation}, we provide qualitative visualizations for our decoder ablation on degradation-free image generation. To explicitly evaluate the effectiveness of our decoupling strategy, we compare the decoding of the exact same scene latent—generated during WaterGen's latent diffusion stage—using two distinct decoders: a vanilla VAE decoder, and our proposed medium-conditional decoder parameterized with strictly degradation-free water conditions ($B^\infty=0, \beta=0$). The visual results clearly demonstrate that WaterGen's medium-conditional decoding yields significantly cleaner and genuinely degradation-free underwater scenes compared to the vanilla VAE baseline. This performance not only highlights the precision of our medium-conditional decoding but also provides strong lateral evidence validating our core contribution: the successful and complete decoupling of scene content from medium degradation.

\section{More Visualizations of Medium Control}

\begin{figure}[h]
    \centering
    \includegraphics[width=\linewidth]{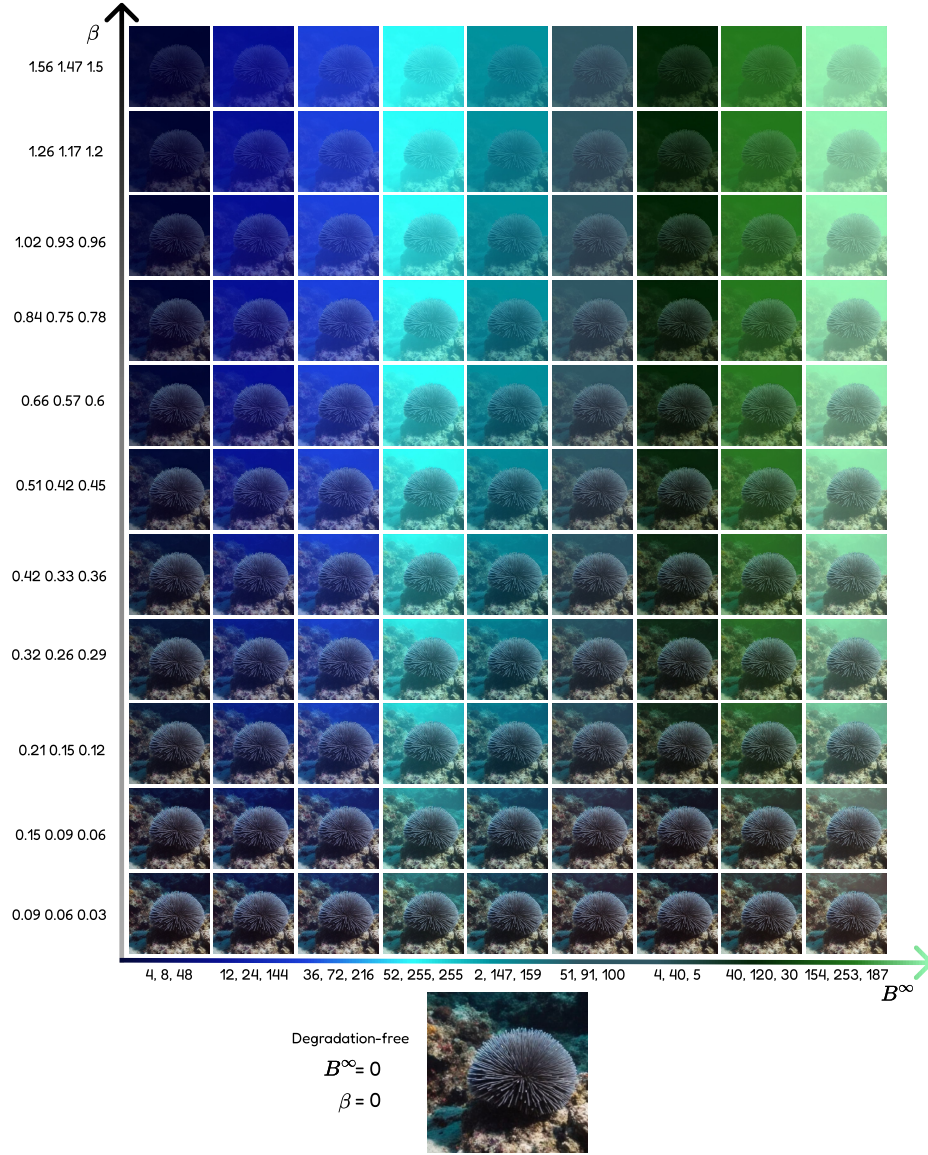}
    \caption{\textbf{Precise medium degradation synthesis.} We fix a single clean scene latent and vary the physical medium parameters to visualize how WaterGen independently controls underwater appearance. Columns change the background light $B^\infty$ (shown as RGB triplets at the bottom), while rows vary the attenuation coefficients $\beta$. The bottom image shows the degradation-free reference. WaterGen produces smooth and physically consistent transitions across both axes, highlighting that our decoder enables fine-grained medium manipulation while preserving the underlying scene content.}
    \label{fig:media_control}
\end{figure}

In \cref{fig:media_control}, we provide additional visualizations to illustrate the fine-grained and independent medium control enabled by WaterGen. Starting from a single degradation-free scene latent, we decode the same underlying content under a grid of water conditions obtained by varying the physically meaningful medium parameters. Specifically, the columns correspond to different background light colors $B^\infty$, while the rows vary the attenuation coefficients $\beta$. The degradation-free reference is shown separately at the bottom.

We show that WaterGen yields smooth and predictable appearance changes as the medium parameters vary. Importantly, these changes are achieved while preserving the same scene geometry and object layout, confirming that medium effects are controlled independently from scene synthesis.
These visualizations further support our core idea: by decoupling semantic scene generation from medium-conditioned decoding, the model supports “one scene, many waters,” enabling precise re-rendering of a fixed scene under diverse optical conditions. This property is especially useful for generating aligned clean/degraded training pairs and for systematically studying the effect of individual medium parameters on downstream underwater vision tasks.

\section{Depth and Transmission Map Visualizations}
\begin{figure}[t]
    \centering
    \includegraphics[width=\linewidth]{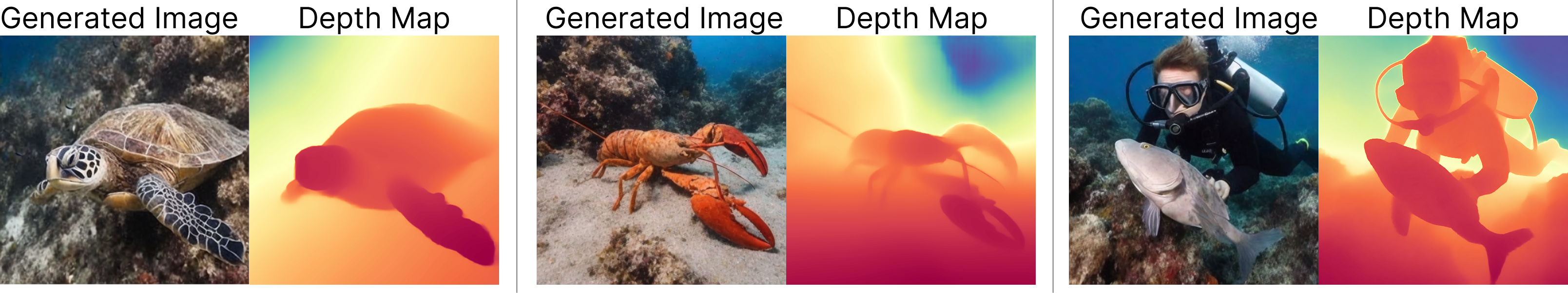}
    \caption{\textbf{Depth-map visualizations for generated scenes.} Generated underwater images are paired with spatially consistent depth maps, supporting physically meaningful construction of degradation maps for conditional decoding.}
    \label{fig:supp_depth_maps}
\end{figure}

In \cref{fig:supp_depth_maps}, we show depth-map visualizations that are spatially consistent with generated scenes, supporting physically meaningful B/T map generation.

\section{Restoration Visualization Results}
\begin{figure}[h]
    \centering
    \includegraphics[width=\linewidth]{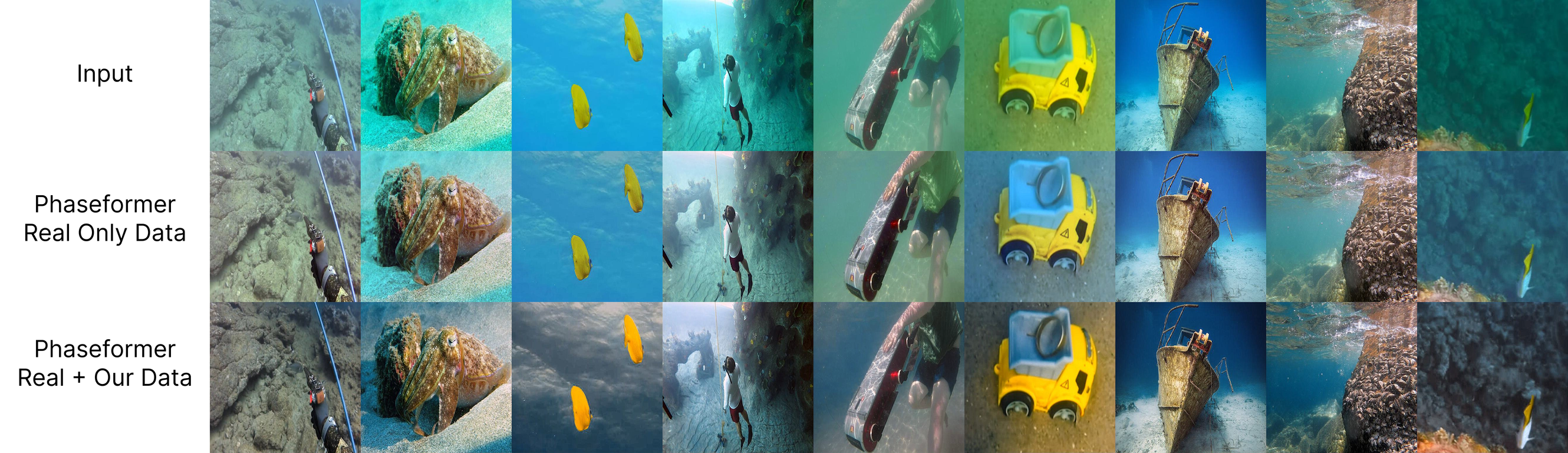}
    \caption{\textbf{Qualitative restoration comparison.} From top to bottom: input underwater images, Phaseformer~\cite{khan2024phaseformer} trained on real data only, and Phaseformer trained on real data augmented with our generated data. Training with our data produces clearer structure, improved contrast, and more faithful colors across diverse scenes.}
    \label{fig:qual_restore_comparison}
\end{figure}

In Fig.~\ref{fig:qual_restore_comparison} we evaluate the impact of our generated data when used to augment real underwater restoration training. We train the same baseline restoration model (Phaseformer~\cite{khan2024phaseformer}) under two settings: (i) using only real underwater training data (\emph{Real Only}), and (ii) using the same real data augmented with our generated samples (\emph{Real + Our Data}). All training hyperparameters and inference settings are kept identical across the two models to isolate the effect of data augmentation. As shown in Fig.~\ref{fig:qual_restore_comparison}, augmenting with our data consistently improves perceptual restoration quality across diverse scenes (e.g., open-water fish, diver, and near-field objects). In particular, the \emph{Real + Our Data} model reduces haze and color cast, recovers higher local contrast, and restores finer textures (e.g., seabed and object boundaries) while maintaining scene structure. These qualitative gains indicate improved generalization to challenging real-world degradations and are consistent with the metric improvements reported in the main paper (\cref{tab:uiis10k_uiqm_musiq}).

\begin{figure}[t]
    \centering
    \includegraphics[width=\linewidth]{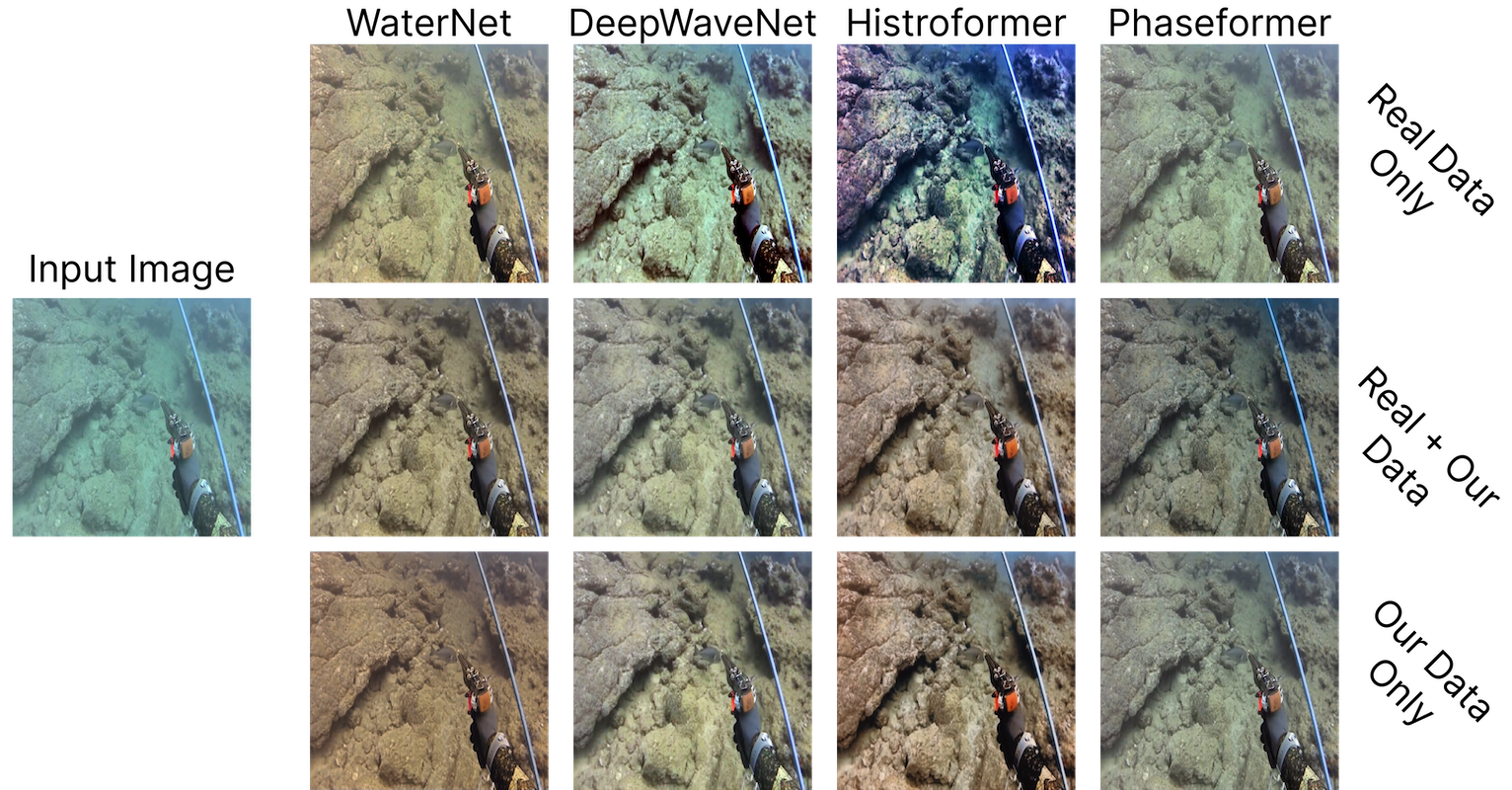}
    \caption{\textbf{Additional restoration comparisons across training settings.} We compare restoration outputs for models trained with real data only, real data augmented with WaterGen data, and WaterGen data only. Augmenting with our generated samples improves contrast, suppresses residual haze, and better preserves local scene details across the tested restoration backbones.}
    \label{fig:supp_restoration_backbones}
\end{figure}

We further show qualitative comparisons for all tested restoration backbones, including Real Data Only, Real+Our Data, and Ours Only training in \cref{fig:supp_restoration_backbones}. These results demonstrate consistent visual improvements across restoration architectures.

\section{Segmentation Visualization Results}
\begin{figure}[t]
    \centering
    \includegraphics[width=\linewidth]{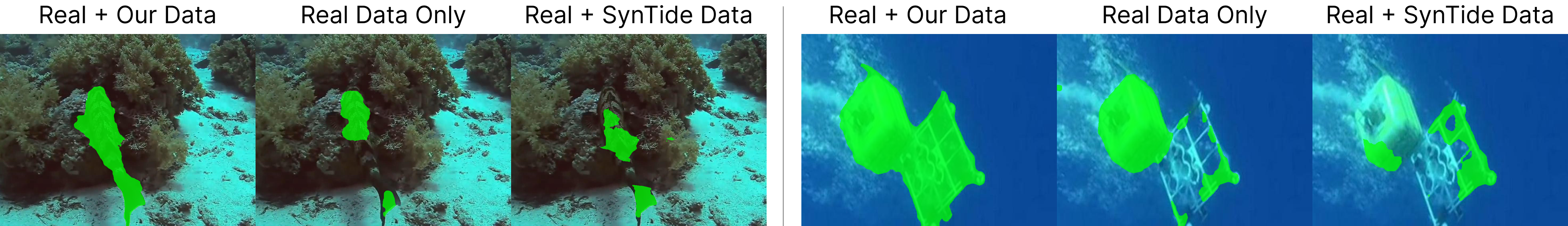}
    \caption{\textbf{Qualitative segmentation comparison.} SegFormer trained with Real+Ours produces more complete object masks under strong underwater degradation, with fewer missed regions and sharper boundaries than the compared training settings.}
    \label{fig:supp_segmentation_comparison}
\end{figure}

In \cref{fig:supp_segmentation_comparison}, we show Real+SynTIDE qualitative results using the SegFormer backbone. Training with Real+Ours produces more complete masks in strongly degraded underwater scenes.

\section{More Details on Ablation Study}
The background light extraction process follows the ULAP~\cite{song2018rapid} framework. Specifically, we first identify a candidate set comprising the top 0.1\% of pixels corresponding to the largest values in the scene depth map. The global background light vector is subsequently determined by selecting the specific pixel within this candidate set that exhibits the maximum $L_2$ norm (i.e., the highest intensity magnitude) in the original RGB color space.

\begin{figure}[t]
    \centering
    \includegraphics[width=\linewidth]{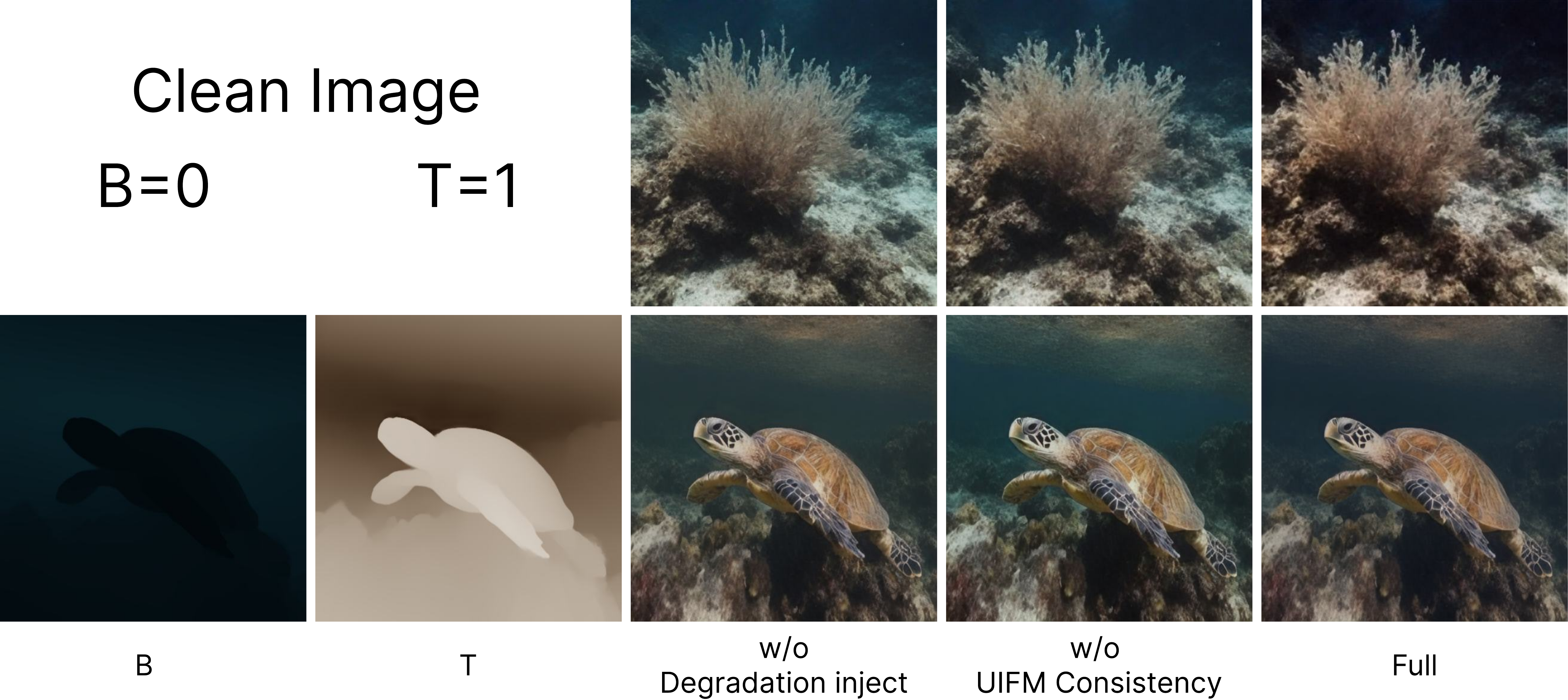}
    \caption{\textbf{Visual ablation of internal training components.} Removing stochastic degradation injection or bidirectional UIFM consistency weakens medium-color accuracy and introduces residual artifacts. The full model produces more faithful medium appearance while preserving the underlying scene.}
    \label{fig:supp_component_ablation}
\end{figure}

We show the visualizations for Degradation Inject and Bidir UIFM Consistency Loss ablations in \cref{fig:supp_component_ablation}. Both components mitigate residual medium degradation entangled in the generated latents, which otherwise interferes with physically accurate conditional decoding. Removing either component reduces underwater color accuracy due to cascaded artifacts.

\end{document}